\documentclass{article}

\usepackage{microtype}
\usepackage{graphicx,float}
\usepackage[utf8]{inputenc}
\usepackage{subfig}
\usepackage{subfig}
\usepackage{bm, color}
\usepackage{amsthm}
\usepackage{amsmath}
\usepackage{booktabs} 

\usepackage{hyperref}

\newcommand\p{\bm{p}}
\newcommand\x{\bm{x}}
\newcommand\y{\bm{y}}
\newcommand\w{\bm{w}}
\newcommand\dx{\mathrm{d}}
\newcommand\B{\mathrm{Beta}}

\usepackage[accepted]{icml2019}

\begin{document}

\twocolumn[
\icmltitle{Quantifying Intrinsic Uncertainty in Classification via Deep Dirichlet Mixture Networks}




\begin{icmlauthorlist}
\icmlauthor{Qingyang Wu}{ucd}
\icmlauthor{He Li}{nyu}
\icmlauthor{Lexin Li}{ucb}
\icmlauthor{Zhou Yu}{ucd}
\end{icmlauthorlist}

\icmlaffiliation{ucd}{Computer Science Department,
University of California, Davis}
\icmlaffiliation{ucb}{Department of Biostatistics and Epidemiology, University of California, Berkeley}
\icmlaffiliation{nyu}{Stern School of Business, New York University}

\icmlcorrespondingauthor{}{}

\icmlkeywords{Machine Learning, ICML}

\vskip 0.3in
]



\printAffiliationsAndNotice{}  

\begin{abstract}
With the widespread success of deep neural networks in science and technology, it is becoming increasingly important to quantify the uncertainty of the predictions produced by deep learning. In this paper, we introduce a new method that attaches an explicit uncertainty statement to the probabilities of classification using deep neural networks. Precisely, we view that the classification probabilities are sampled from an unknown distribution, and we propose to learn this distribution through the Dirichlet mixture that is flexible enough for approximating any continuous distribution on the simplex. We then construct credible intervals from the learned distribution to assess the uncertainty of the classification probabilities. Our approach is easy to implement, computationally efficient, and can be coupled with any deep neural network architecture. Our method leverages the crucial observation that, in many classification applications such as medical diagnosis, more than one class labels are available for each observational unit. We demonstrate the usefulness of our approach through simulations and a real data example.
\end{abstract}

\section{Introduction}

Deep neural networks have been achieving remarkable success in a wide range of classification tasks in recent years. Accompanying increasingly accurate prediction of the classification probability, it is of equal importance to quantify the uncertainty of the classification probability produced by deep neural networks. Without a careful characterization of such an uncertainty, the prediction of deep neural networks can be questionable, unusable, and in the extreme case incur considerable loss \cite{wang2016deep}. For example, deep reinforcement learning suffers from a strikingly low reproducibility due to high uncertainty of the predictions \cite{henderson2017deep}. Uncertainty quantification can be challenging though; for instance, \cite{guo2017calibration} argued that modern neural networks architectures are poor in producing well-calibrated probability in binary classification. Recognizing such challenges, there have been recent proposals to estimate and quantify the uncertainty of output from deep neural networks, and we review those methods in Section \ref{sec:related}. Despite the progress, however, uncertainty quantification of deep neural networks remains relatively underdeveloped \cite{kendall2017uncertainties}. 

In this paper, we propose deep Dirichlet mixture networks to produce, in addition to a point estimator of the classification probabilities, an associated credible interval (region) that covers the true probabilities at a desired level. We begin with the binary classification problem and employ the Beta mixture model to approximate the probability distribution of the true but \text{random} probability. We then extend to the general multi-class classification using the Dirichlet mixture model. Our key idea is to view the classification probability as a random quantity, rather than a deterministic value in $[0,1]$. We seek to estimate the distribution of this random quantity using the Beta or the Dirichlet mixture, which we show is flexible enough to model any continuous distribution on $[0,1]$. We achieve the estimation by adding an extra layer in a typical deep neural network architecture, without having to substantially modify the overall structure of the network. Then based on the estimated distribution, we produce both a point estimate and a credible interval for the classification probability. This credible interval provides an explicit quantification of the classification variability, and can greatly facilitate our decision making. For instance, a point estimate of high probability to have a disease may be regarded as lack of confidence if the corresponding credible interval is wide. By contrast, a point estimate with a narrow credible interval may be seen as a more convincing  diagnosis. 

The feasibility of our proposal is built upon a crucial observation that, in many classification applications such as medical diagnosis, there exist more than one class labels. For instance, a patient's computed tomography image may be evaluated by two doctors, each giving a binary diagnosis of existence of cancer. In Section \ref{sec:realdata}, we illustrate with an example of diagnosis of Alzheimer's disease (AD) using patients' anatomical magnetic resonance imaging. For each patient, there is a binary diagnosis status as AD or healthy control, along with additional cognitive scores that are strongly correlated with and carry crucial information about one's AD status. We thus consider the dichotomized version of the cognitive scores, combine them with the diagnosis status, and feed them together into our deep Dirichlet mixture networks to obtain a credible interval of the classification probability. We remark that, existence of multiple labels is common rather than an exception in a variety of real world applications.

Our proposal provides a useful addition to the essential yet currently still growing inferential machinery to deep neural networks learning. Our method is simple, fast, effective, and can couple with any existing deep neural network structure. In particular, it adopts a frequentist inference perspective, but produces a Bayesian-style outcome of credible intervals.

\subsection{Related Work}
\label{sec:related}


There has been development of uncertainty quantification of artificial neural networks since two decades ago. Early examples include the delta method \cite{hwang1997prediction}, and the Bootstrap methods  \cite{efron1994introduction,heskes1997practical,carney1999confidence}. However, the former requires computing the Hessian matrix and is computationally expensive, whereas the latter hinges on an unbiased prediction. When the prediction is biased, the total variance is to be underestimated, which would in turn result in a narrower credible interval.

Another important line of research is Bayesian neural networks \cite{mackay1992evidence, mackay1992practical}, which treat model parameters as distributions, and thus can produce an explicit uncertainty quantification in addition to a point estimate. The main drawback is the prohibitive computational cost of running MCMC algorithms. There have been some recent proposals aiming to address this issue, most notably, \cite{gal2016dropout, li2017dropout} that used the dropout tricks. Our proposal, however, is a frequentist solution, and thus we have chosen not to numerically compare with those Bayesian approaches. 

Another widely used uncertainty quantification method is the mean variance estimation (MVE) approach  \cite{nix1994estimating}. It models the data noise using a normal distribution, and employs a neural network to output the mean and variance. The optimization is done by minimizing the negative log-likelihood function. It has mainly been designed for regression tasks, and is less suitable for classification.

There are some more recent proposals of uncertainty quantification. One is the lower and upper bound estimation (LUBE) \cite{khosravi2011lower,quan2014short}. LUBE has been proven successful in numerous applications. However, its loss function is non-differentiable and gradient descent cannot be applied for optimization. The quality-driven prediction interval method (QD) has recently been proposed to improve LUBE \cite{pearce2018high}. It is a distribution-free method by outputting prediction's upper bound and lower bound. The uncertainty can be estimated by measuring the distance between the two bounds. Unlike LUBE, the objective function of QD can be optimized by gradient descent. But similar to MVE, it is designed for regression tasks. Confidence network is another method to  estimate confidence by adding an output node next to the softmax probabilities \cite{devries2018confidence}. This method is suitable for classification. Although its original goal was for out-of-distribution detection, its confidence can be used to represent the intrinsic uncertainty. Later in Section \ref{sec:compare}, we numerically compare our method with MVE, QD, and confidence network. 

We also clarify that our proposed framework is different from the mixture density network \cite{bishop1994mixture}. The latter trains a neural network to model the distribution of the outcome using a mixture distribution. By contrast, we aim to learn the distribution of the classification probabilities and to quantify their variations.

\section{Dirichlet Mixture Networks}

In this section, we describe our proposed Dirichlet mixture networks. We begin with the case of binary classification, where the Dirichlet mixture models reduce to the simpler Beta mixture models. Although a simpler case, the binary classification is sufficient to capture all the key ingredients of our general approach and thus loses no generality. At the end of this section, we discuss the extension to the multi-class case.

\subsection{Loss Function}

\begin{figure}[t!]
\centering
\includegraphics[width=8.3cm]{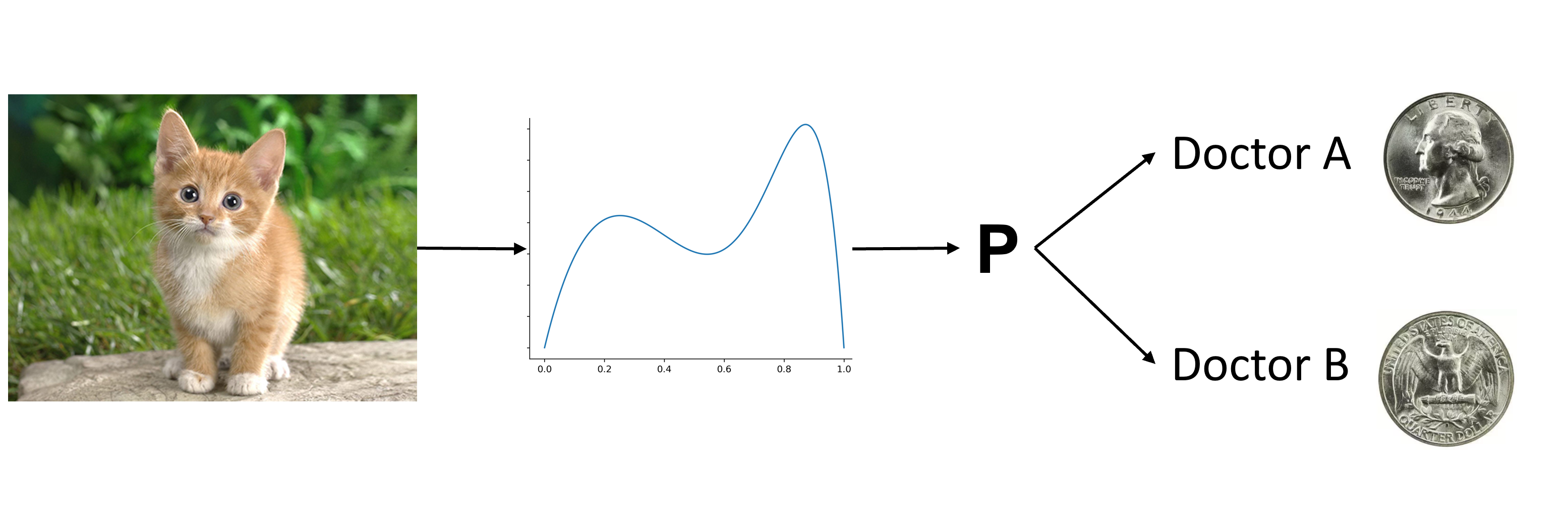}
\caption{Illustration of the setting. Two or more labels are generated with the \textit{same} probability $p_{\x}$, which is \textit{randomly} drawn from a distribution that we wish to estimate.}
\label{fig:illlustration}
\end{figure}

We begin with a description of the key idea of our proposal. Let $\{1, 2\}$ denote the two classes. Given an observational unit $\x$, e.g., an image, we view the probability $p_{\x}$ that $\x$ belongs to class 1 as a random variable, instead of a deterministic value in $[0, 1]$. We then seek to estimate the probability density function $f(p; \x)$ of $p_{\x}$. This function encodes the \textit{intrinsic} uncertainty of the classification problem. A point estimate of the classification probability only focuses on the mean, $\int_0^1 f(p; \x) \dx p$, which is not sufficient for an informed decision making without an explicit quantification of its variability. For example, it can happen that, for two observational units $\x$ and $\x'$, their mean probabilities, and thus their point estimates of the classification probability, are the same. However, the densities are far apart from each other, leading to completely different variabilities, and different interpretations of the classification results. Figure \ref{fig:illlustration} shows an illustration. Our proposal then seeks to estimate the density function $f(p; \x)$ for each $\x$.

A difficulty arising from this estimation problem is that $f$ in general can be any density function on $[0, 1]$. To address this, we propose to simplify the problem by restricting to the case where $f$ is a Beta mixture; i.e., 
\begin{equation}\label{eq:beta_density_mix}
f(p; \x) = \sum_{k=1}^K w^k \frac{p^{\alpha^k_1 -1} (1-p)^{\alpha^k_2 -1}}{\B(\alpha^k_1, \alpha^k_2)},
\end{equation}
where $\B(\cdot, \cdot)$ is the Beta function, and the parameters $w^k, \bm\alpha^k = (\alpha^k_1, \alpha^k_2)$ are smooth functions of $\x$, $k = 1, \ldots, K$. The weights $w^k$ satisfy that $w^1 + \cdots + w^K = 1$. Later we show that this Beta mixture distribution is flexible enough to adequately model almost any distribution $f$ on $[0,1]$. 

With the form of density function \eqref{eq:beta_density_mix} in place, our goal then turns to estimate the positive parameters $\alpha_1^k, \alpha_2^k$, and $w^k$. To do so, we derive the loss function that is to be minimized by deep neural networks.

We employ the negative log-likelihood function from \eqref{eq:beta_density_mix} as the loss function. For the $j$th observational unit of the training data, $j = 1, \ldots, n$, let $\x_j$ denote the input, e.g., the subject's image scan, and $\y_j = \left( y_j^{(1)}, \ldots, y_j^{(m_j)} \right)$ denote the vector of labels taking values from $\{1, 2\}$. Here we assume $m_j \geq 2$, reflecting that there are more than one class label for each observational unit. Write $\w = (w_1, \ldots, w_K)$ and $\bm \alpha = (\bm\alpha^1, \ldots, \bm\alpha^K)$. By integrating out $p$, the likelihood function for the observed pair $(\x_j, \y_j)$ is 
\[
\begin{aligned}
&L_j(\w, \bm\alpha; \x_j, \y_j) \\
&= \int_0^1 p^{\sum_{l=1}^{m_j} \bm{1}(y_j^{(l)}= 1)} (1-p)^{\sum_{l=1}^{m_j} \bm{1}(y_j^{(l)}=2)} f(p; \x_j)  \dx p.
\end{aligned}
\]
Write $S_{ij} = \sum_{l=1}^{m_j} \bm{1} \left( y_j^{(l)} = i \right)$, where $\bm{1}(\cdot)$ is the indicator function, $i = 1, 2$, $j = 1, \ldots, n$, and this term quantifies the number of times $\x_j$ is labeled $i$. Plugging \eqref{eq:beta_density_mix} into $L_j$, we get 
\[
\begin{aligned}
& L_j(\w, \bm\alpha; \x_j, \y_j) \\
&= \int_0^1 p^{S_{1j}} (1-p)^{S_{2j}} \sum_{k=1}^K \frac{w^k p^{\alpha^k_1 -1} (1-p)^{\alpha^k_2 -1}}{\B(\alpha^k_1, \alpha^k_2)} \dx p\\
&= \sum_{k=1}^K \int_0^1 \frac{w^k}{\B(\alpha^k_1, \alpha^k_2)}  p^{\alpha^k_1 -1 + S_{1j}} (1-p)^{\alpha^k_2 -1 + S_{2j}} \dx p.
\end{aligned}
\]
By a basic property of Beta functions, we further get
\[
L_j(\w, \bm\alpha; x_j, \bm y_j) = \sum_{k=1}^K \frac{w^k\B(\alpha^k_1  + S_{1j}, \alpha^k_2 + S_{2j})}{\B(\alpha^k_1, \alpha^k_2)}.
\]
Aggregating all $n$ observational units, we obtain the full negative log-likelihood function, 
\begin{equation}\label{eq:full_like}
\begin{aligned}
& -\ell(\w, \bm\alpha;  \x_1, \y_1, \ldots, \x_n, \y_n) \\
& = -\sum_{j=1}^n \log \left[\sum_{k=1}^K \frac{w^k\B(\alpha^k_1  + S_{1j}, \alpha^k_2 + S_{2j})}{\B(\alpha^k_1, \alpha^k_2)} \right].
\end{aligned}
\end{equation}

We then propose to employ a deep neural network learner to estimate $\w$ and $\bm \alpha$.

\subsection{Credible Intervals}
\label{sec:credible-interval}

To train our model, we simply replace the existing loss function of a deep neural network, e.g., the cross-entropy, with the negative log-likelihood function given in (2). Therefore, we can take advantage of current deep learning framework such as PyTorch for automatic gradient calculation. Then we use the mini-batch gradient descent to optimize the entire neural network’s weights. Once the training is finished, we obtain the estimate of the parameters of the mixture distribution, $\{\bm{w}, \bm{\alpha}\}$.

One implementation detail to notice is that the Beta function has no closed form derivative. To address this issue, we used fast log gamma algorithm to obtain an approximation of the Beta function, which is available in PyTorch. Also, we applied the softmax function to the weights of the mixtures to ensure that $w_1 + ... + w_K = 1$, and took the exponential of $\bm{\alpha}^1, \ldots \bm{\alpha}^K$ to ensure that these parameters remain positive as required. 

Given the estimated parameters $\widehat{\w}, \widehat{\bm \alpha}$ from the deep mixture networks, we next construct the credible interval for explicit uncertainty quantification. For a new observation $\x_0$, the estimated distribution of the classification probability $p_{\x_0}$ takes the form
\[
\hat f(p; \x_0) = \sum_{k=1}^K \hat w^k(\x_0) \frac{p^{\hat\alpha^k_1(\x_0) -1} (1-p)^{\hat\alpha^k_2(\x_0) -1}}{\B(\hat\alpha^k_1(\x_0), \hat\alpha^k_2(\x_0))},
\]
where we write $\hat{w}^k, \hat{\alpha}^k_1, \hat{\alpha}^k_2$ in the form of explicit functions of $\x_0$. The expectation of this estimated density $\int_0^1 \hat f(p; \x_0) \dx p$ is an approximately unbiased estimator of $p_{\x_0}$. Meanwhile, we can construct the two-sided credible interval of $p_{\x_0}$ with the nominal level $\alpha \in (0,1)$ as  
\[
\left[ \widehat Q_{\frac{\alpha}{2}}, \widehat Q_{1 - \frac{\alpha}{2}} \right],
\]
where $\widehat Q_{\frac{\alpha}{2}}$ and $\widehat Q_{1 - \frac{\alpha}{2}}$ are the $\alpha/2$ and $1 - \alpha/2$ quantiles of the estimated density $\hat{f}(p; \x_0)$. Similarly, we can construct the upper and lower credible intervals as 
\[
\left[0,  \widehat Q_{1 - \alpha}\right], \text{ and } \left[\widehat Q_{\alpha}, 1\right],
\]
respectively, where $\widehat Q_{\alpha}$ and $\widehat Q_{1 - \alpha}$ are the $\alpha$ and $1 - \alpha$ quantiles of the estimated density $\hat{f}(p; \x_0)$. 

Next we justify our choice of Beta mixture for the distribution of classification probability, by showing that any density function under certain regularity conditions can be approximated well by a Beta mixture. Specifically, denote by $\mathcal P$ the set of all probability density functions $f$ on $[0, 1]$ with at most countable discontinuities that satisfy
\begin{align}\nonumber
\int_0^1 f(p) \left|\log f(p)\right| \dx p < \infty.
\end{align}
It is shown in \cite{article} that any $f \in \mathcal{P}$ can be approximated arbitrarily well by a sequence of Beta mixtures. That is, for any $f \in \mathcal{P}$ and any $\epsilon > 0$, there exists a Beta mixture distribution $f_{\B}$ such that
\begin{align}\nonumber
\mathrm{D}_{\mathrm{KL}} \left(f \| f_{\B} \right) \leq \epsilon,
\end{align}
where $\mathrm{D}_{\mathrm{KL}} (\cdot \| \cdot)$ denotes the Kullback-Leibler divergence. This result establishes the validity of approximating a general distribution function using a Beta mixture. The proof of this result starts by recognizing that $f$ can be accurately approximated by piecewise constant functions on $[0,1]$ due to a countable number of discontinuities. Next, each constant piece is a limit of a sequence of Bernstein polynomials, which are infinite Beta mixtures with integer parameters \cite{verdinelli1998bayesian, petrone2002consistency}.

\subsection{Multiple-class Classification}
\label{sec:extens-mult-labels}

We next extend our method to the general case of multi-class classification. It follows seamlessly from the prior development except that now the labels $\y_j = \left( y_j^{(1)}, \ldots, y_j^{(m_j)} \right)$ take values from $\{1, 2, \ldots, d\}$, where $d$ is the total number of classes. Given an observation $\x$, the multinomial distribution over $\{1, 2, \ldots, d\}$ is represented by $\p = (p_1, \ldots, p_d)$, which, as a point in the simplex $\Delta = \{(c_1, \ldots, c_d): c_i \ge 0, c_1 + \cdots + c_d = 1\}$, is assumed to follow a Dirichlet mixture
\[
f(\p; \x) = \sum_{k=1}^K w^k \frac1{\B(\bm\alpha^k)} \prod_{i=1}^d p_i^{\alpha^k_i -1}, 
\]
where the generalized Beta function takes the form
\[
\B(\bm\alpha) = \frac{\prod_{i=1}^d \Gamma(\alpha_i)}{\Gamma(\alpha_1 + \cdots + \alpha_d)}.
\]
The likelihood of the $j$th observation is 
\[
L_j = \int_{\Delta} \left( \prod_{i=1}^d p_i^{S_{ij}} \right) \sum_{k=1}^K w^k \frac1{\B (\bm\alpha^k)} \prod_{i=1}^d p_i^{\alpha^k_i -1}  \dx \bm p, 
\]
where $S_{ij} = \sum_{l=1}^{m_j} \bm{1}\left( y_j^{(l)}= i \right)$. Accordingly, the negative log-likelihood function is
\[
\footnotesize
\begin{aligned}
&-\ell(\w, \bm\alpha;  \x_1, \y_1, \ldots, \x_n, \y_n) \\
&= -\sum_{j=1}^n \log \left[\sum_{k=1}^K \frac{w^k\B\left( \alpha^k_1  + S_{1j}, \ldots, \alpha^k_d + S_{dj} \right)}{\B(\bm\alpha^k)} \right].
\end{aligned}
\normalsize
\]
This is the loss function to be minimized in the Dirichlet mixture networks.

\section{Simulations}
\label{sec:simulations}

\subsection{Simulations on Coverage Proportion}

We first investigate the empirical coverage of the proposed credible interval. We used the MNIST handwritten digits data, and converted the ten outcomes (0-9) first to two classes (0-4 as Class 1, and 5-9 as Class 2), then to three classes (0-2 as Class 1, 3-6 as Class 2, and 7-9 as Class 3). In order to create multiple labels for each image, we trained a LeNet-5 \cite{lecun1998gradient} to output the classification probability $p_i$, then sampled multiple labels for the same input image based on a binomial or multinomial distribution with $p_i$ as the parameter. We further divided the simulated data into training and testing sets. We calculated the empirical coverage as the proportion in the testing set that the corresponding $p_i$ falls in the constructed credible interval. We assessed the coverage performance by examining how close the empirical coverage is to the nominal coverage between the interval of 75\% and 95\%. Ideally, the empirical coverage should be the same as the nominal level.

\begin{figure}[h]
\centering
\subfloat[Two labels]{{\includegraphics[width=4cm]{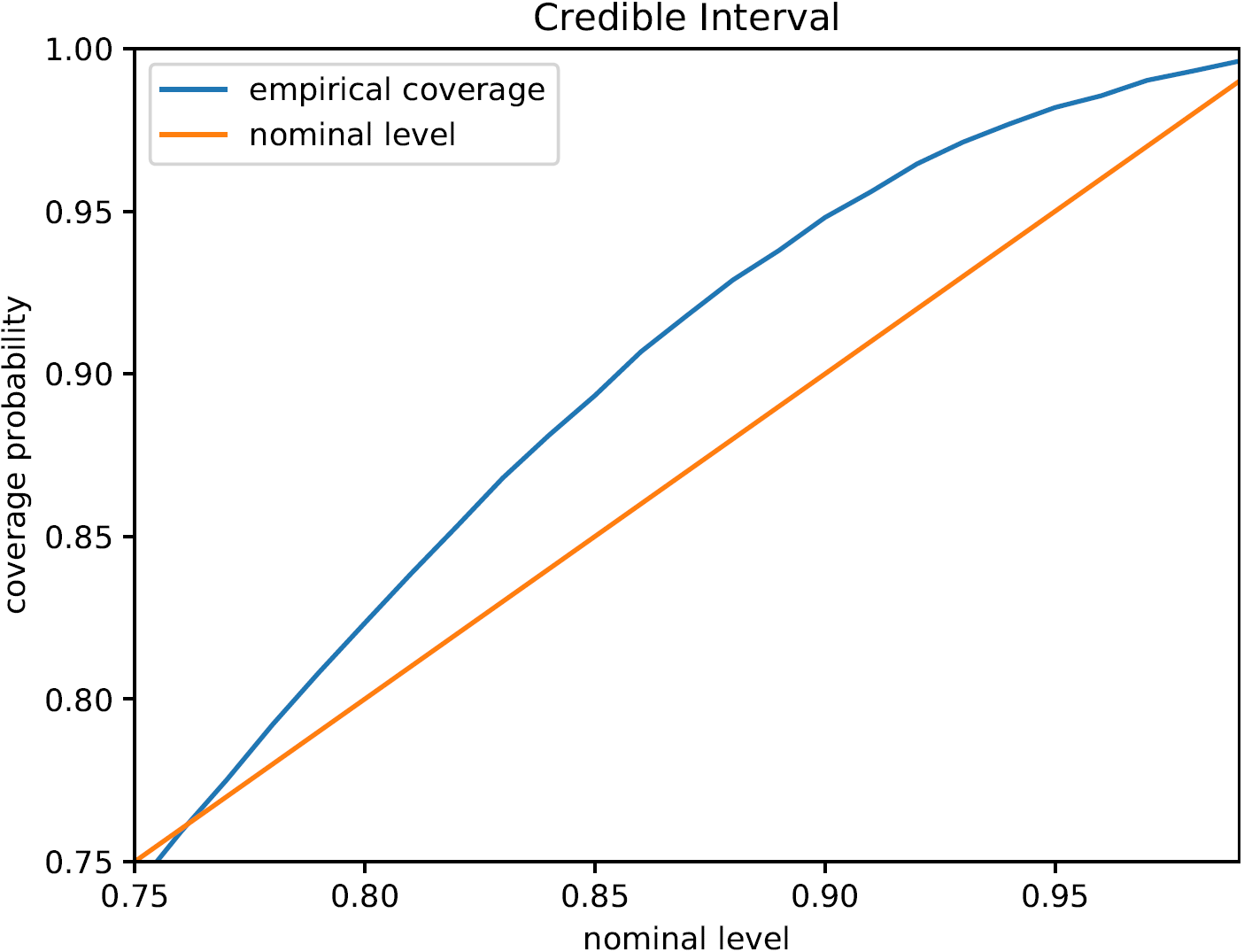}}}
\,
\subfloat[Three labels]{{\includegraphics[width=4cm]{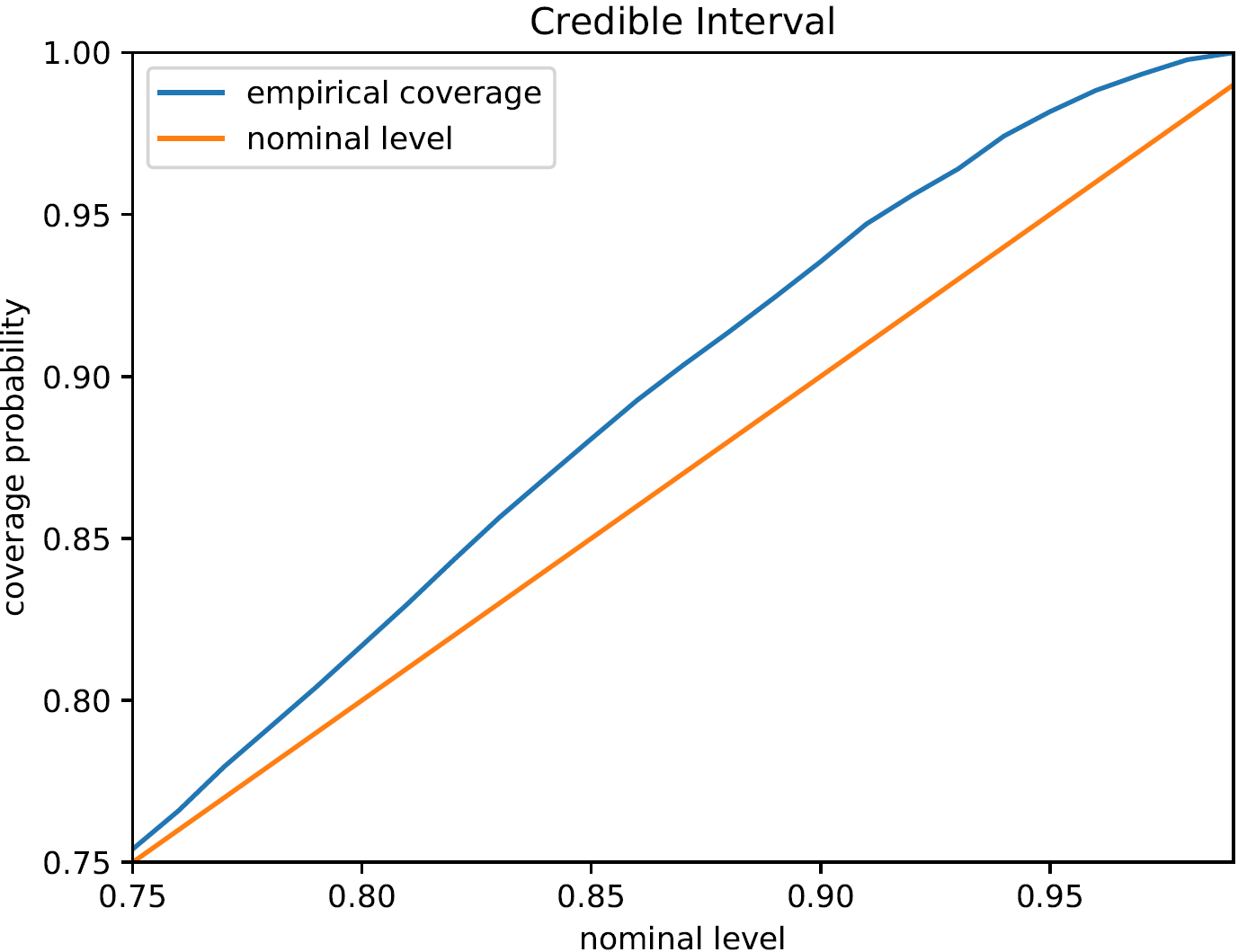}}}
\caption{Empirical coverage of the estimated credible interval for a two-class classification task, with the two-label setting shown in (a), and the three-label setting in (b). The blue line represents the empirical coverage of the estimated credible interval. The orange 45-degree line represents the ideal estimation. The closer the two lines, the better the estimation.}
\label{sim2}
\end{figure}

Figure \ref{sim2} reports the simulation results for the two-class classification task, where panel (a) is when there are two labels available for each input, and panel (b) is when there are three labels available. The orange 45-degree line represents the ideal coverage. The blue line represents the empirical coverage of the credible interval produced by our method. It is seen that our constructed credible interval covers 98.19\% of the truth with the 95\% nominal level for the two-label scenario, and 98.17\% for the three-label scenario. In general, the empirical coverage is close or slightly larger than the nominal value, suggesting that the credible interval is reasonably accurate. Moreover, the interval becomes more accurate with more labels on each input. 

\begin{figure}[h]
\centering
\subfloat[Class 1 with two labels]{{\includegraphics[width=4cm]{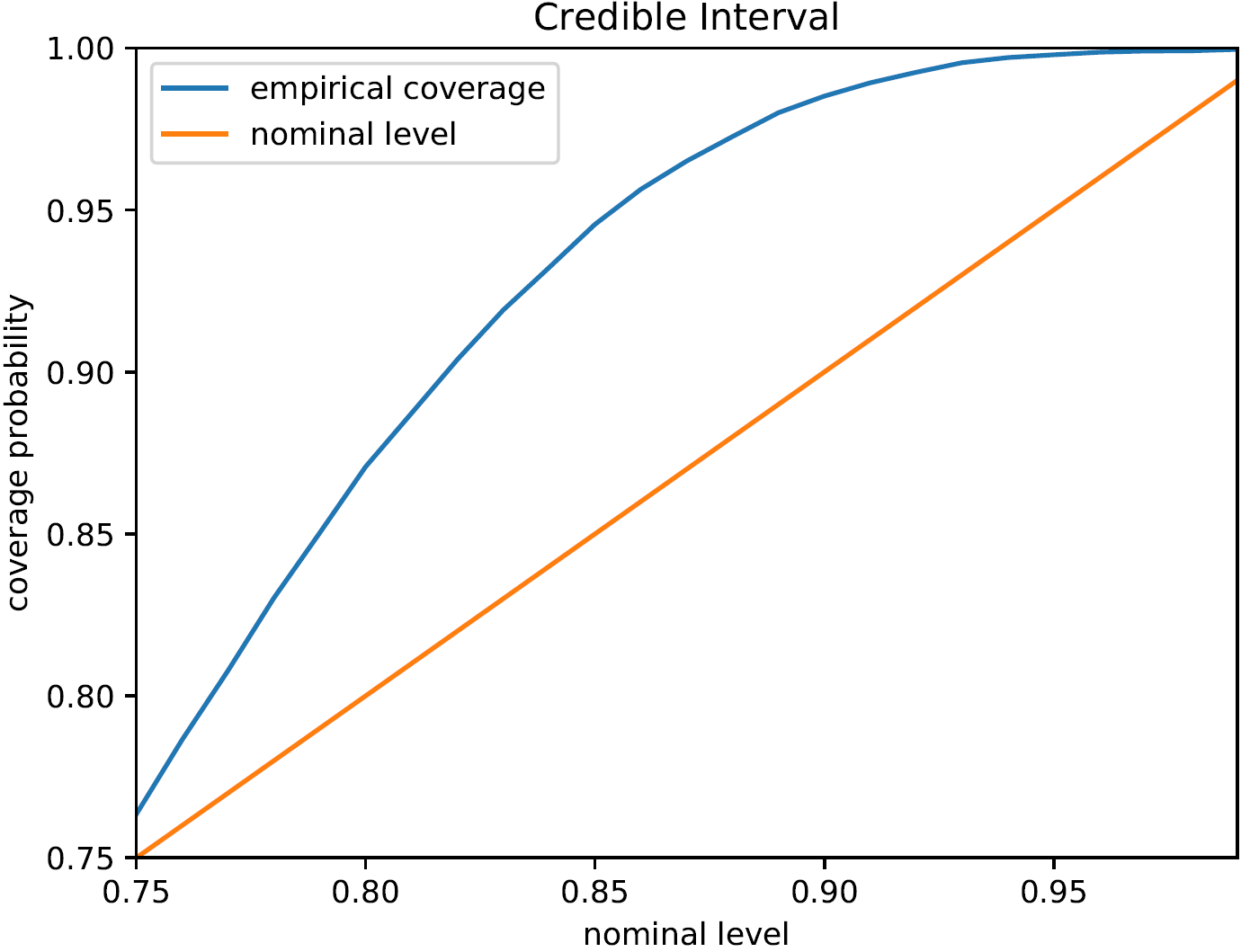}}}
\,
\subfloat[Class 2 with two labels]{{\includegraphics[width=4cm]{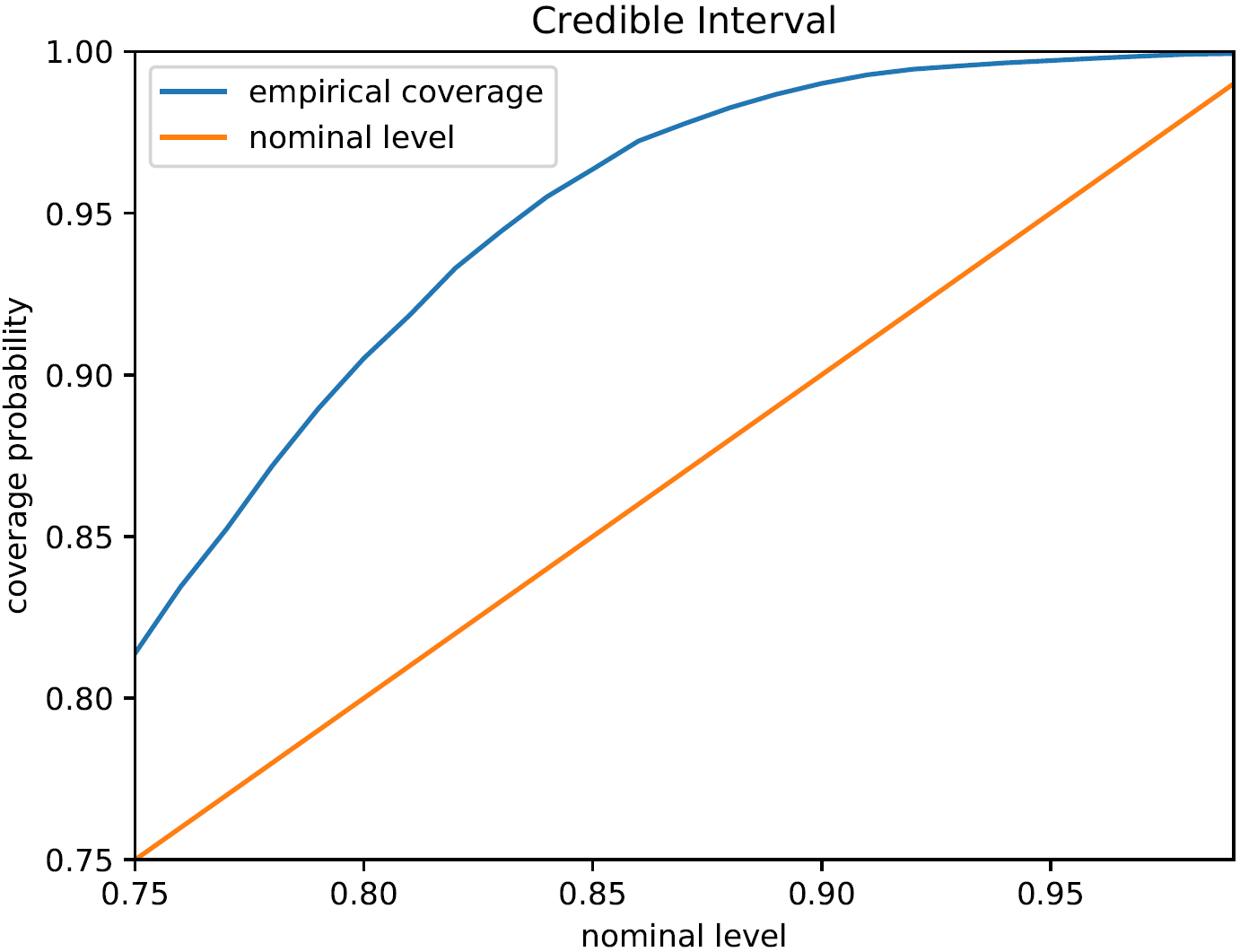}}}
\,
\subfloat[Class 1 with three labels]{{\includegraphics[width=4cm]{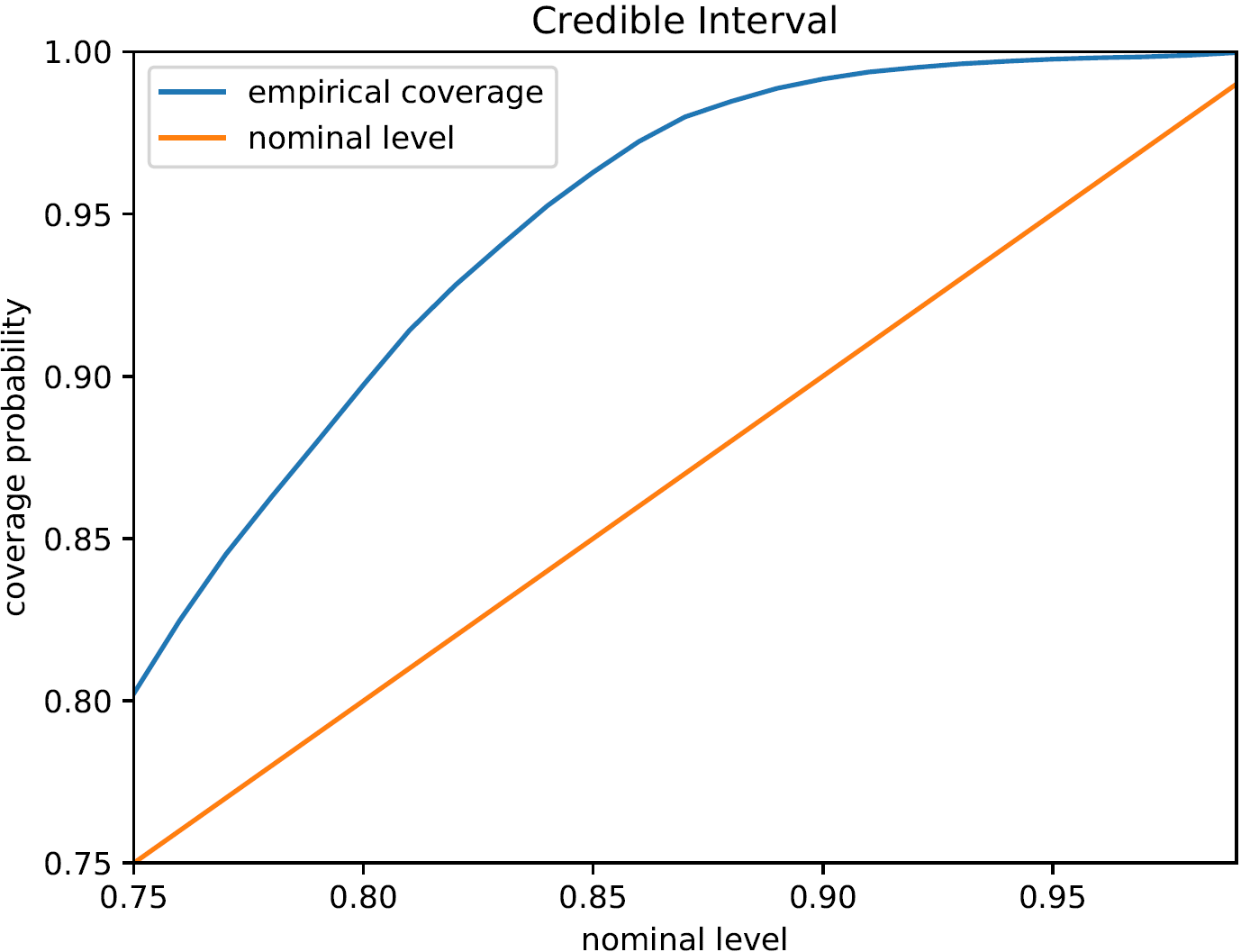}}}
\,
\subfloat[Class 2 with three labels]{{\includegraphics[width=4cm]{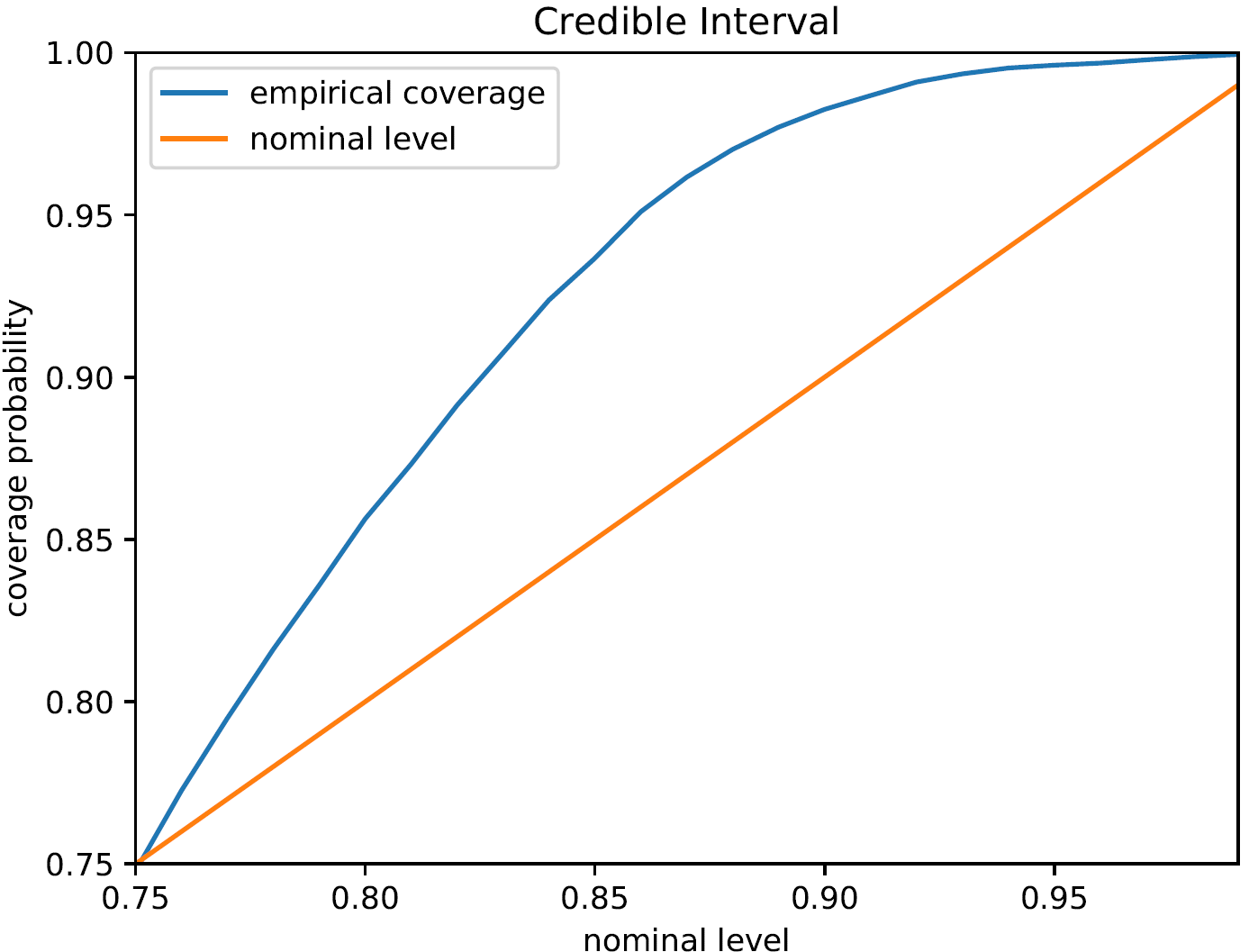}}}
\caption{Empirical coverage of the estimated credible interval for a three-class classification task, with the two-label setting shown in (a) and (b), and the three-label setting in (c) and (d). The blue line represents the empirical coverage of the estimated credible interval. The orange  45-degree line represents the ideal estimation. The closer the two lines, the better the estimation. For each graph, the probability is calculated in the one-vs-all fashion; e.g., (a) represents the credible interval of Class 1 versus Classes 2 and 3 combined.}
\label{sim3}
\end{figure}

Figure \ref{sim3} reports the simulation results for the three-class classification task, where panels (a) and (b) are when there are two labels available, and panels (c) and (d) are when there are three labels available. A similar qualitative pattern is observed in Figure \ref{sim3} as in Figure \ref{sim2}, indicating that our method works well for the three-class classification problem.

\subsection{Comparison with Alternative Methods}
\label{sec:compare}

We next compare our method with three alternatives that serve as the baselines, the confidence network \cite{devries2018confidence}, the mean variance estimation (MVE) \cite{nix1994estimating}, and the quality-driven prediction interval method (QD) \cite{pearce2018high}. We have chosen those methods as baselines, as they also targeted to quantify the intrinsic variability and represented the most recent state-of-the-art solutions to this problem. 

\begin{figure}[h]
    \centering
    \subfloat[Plot for $f_1=\frac{\psi_1}{\psi_2} + 1$]{{\includegraphics[width=4.1cm]{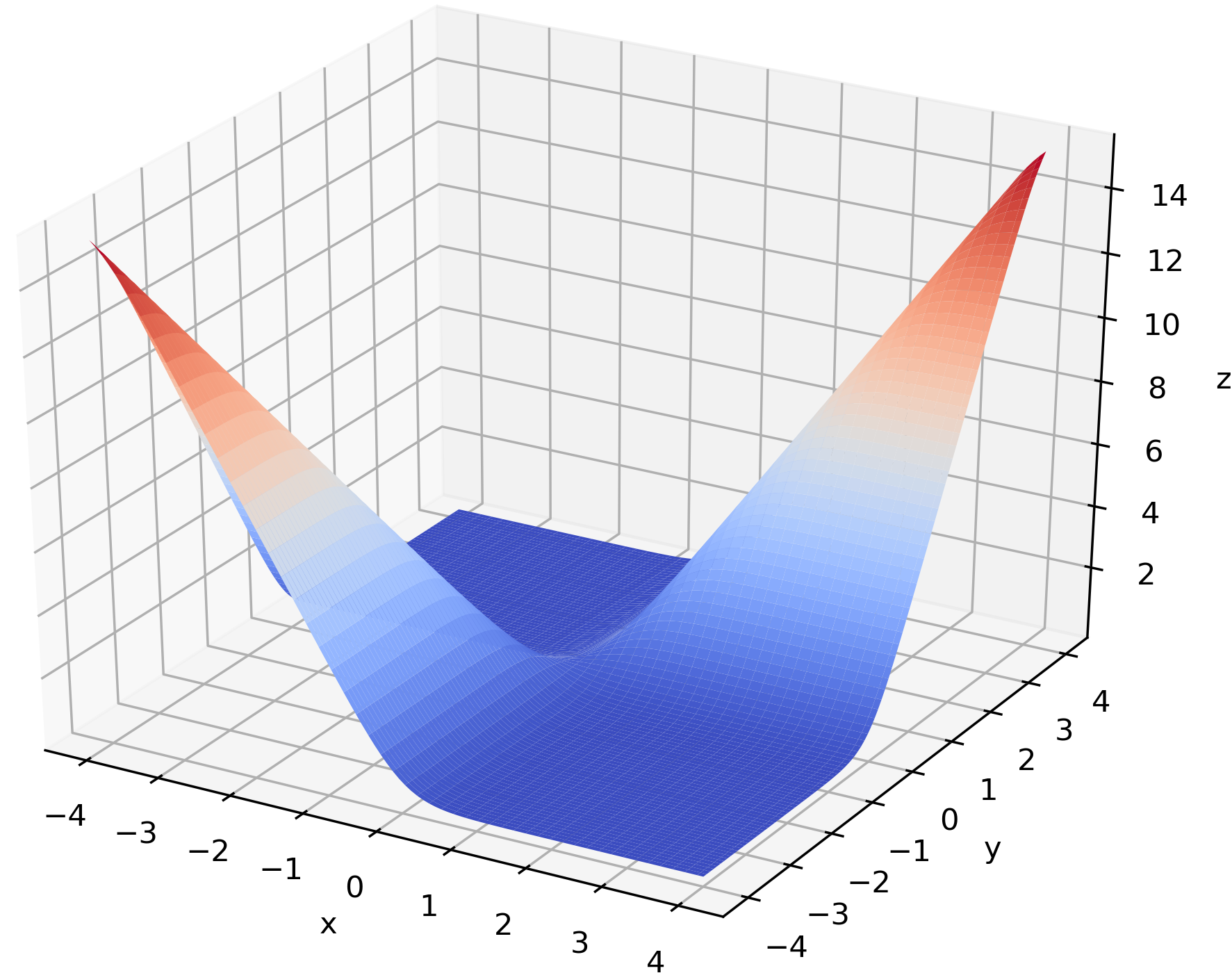}}}
    \subfloat[Plot for $f_2=\frac{\psi_2}{\psi_1} + 1$]{{\includegraphics[width=4.1cm]{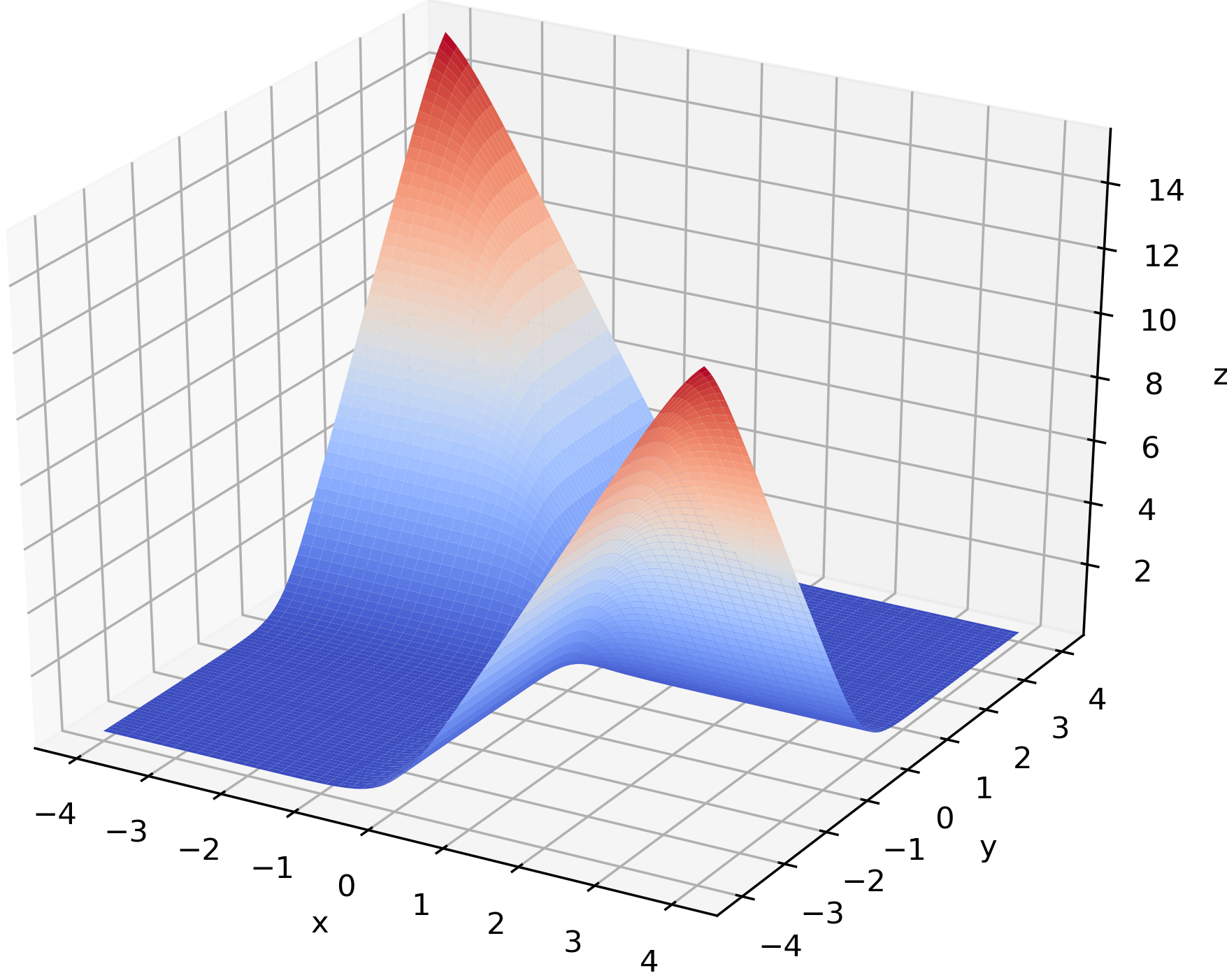}}}
    \,
    \subfloat[Scatter Plot for 1000 samples]{{\includegraphics[width=6cm]{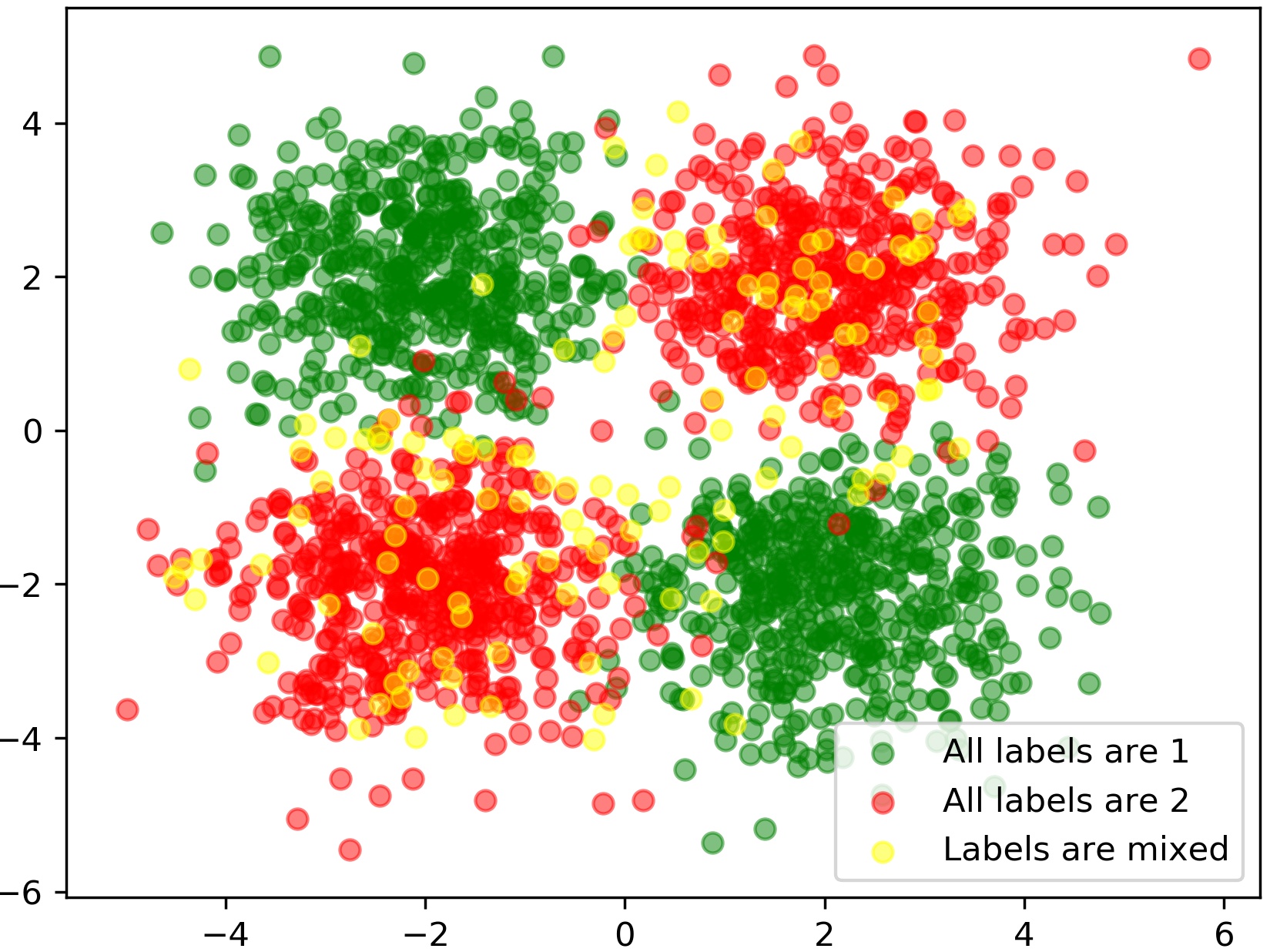}}}
    \caption{Data is generated from a Bernoulli distribution whose parameter is sampled from a $\B$ distribution with parameter($f_1$, $f_2$). (a) and (b) show the 3D landscapes. (c) shows 1,000 samples from this distribution with two labels for each data point. Green means all labels are 1. Red means all labels are 2. Yellow means that labels are a mix of 1 and 2.}
    \label{samples}
\end{figure}

To facilitate graphical presentation of the results, we simulated the input data $\x$ from two-dimensional Gaussian mixtures. Specifically, we first sampled $\x$ from a mixture of two Gaussians with means at $(-2, 2)$ and $(2, -2)$, and denote its probability density function as $\psi_1$. We then sampled $\x$ from another mixture of two Gaussians with means at $(2, 2)$ and $(-2, -2)$, and denote its probability density function as $\psi_2$. For each Gaussian component, the variance is set at 0.7. We then sampled the probability $p$ of belonging to Class 1 from a Beta distribution with the parameters $\psi_1 / \psi_2 + 1$ and $\psi_2 / \psi_1 + 1$. Finally, we sampled the class labels from a Bernoulli distribution with the probability of success $p$. At each input sample $\x$, we sampled two class labels. For a fair comparison, we duplicate the data for the baseline methods that only use one class label. Figure \ref{samples} (c) shows a scatter plot of 1,000 samples, each with two labels. The green dots correspond to the samples whose class labels are 0 in both replications, the red dots  are 1 in both replications, and the yellow dots are those samples whose class labels are different in two replications. Most of the yellow dots are located along the two axis that separate the four quadrants. 

\begin{figure}[h]
    \centering
    \subfloat[Ideal]{{\includegraphics[width=4.5cm]{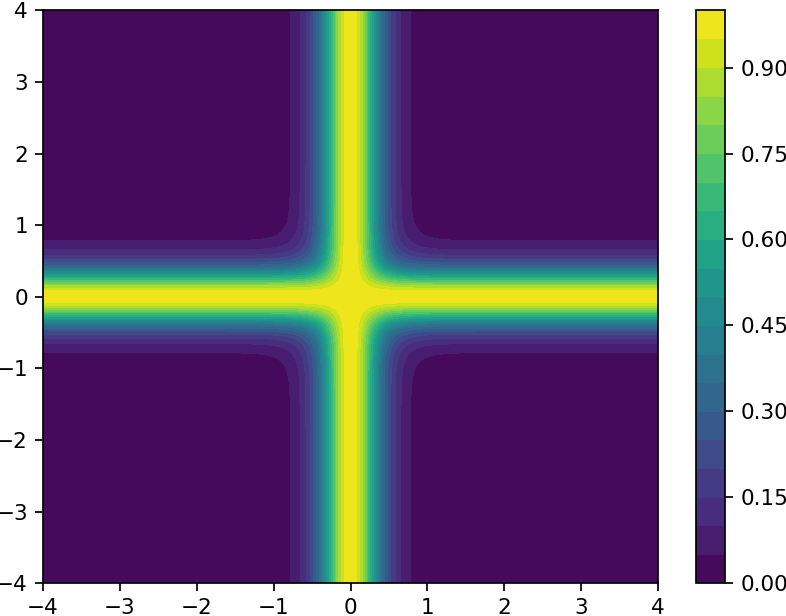}}}
    \hspace{3cm}
    \subfloat[Our Approach]{{\includegraphics[width=4cm]{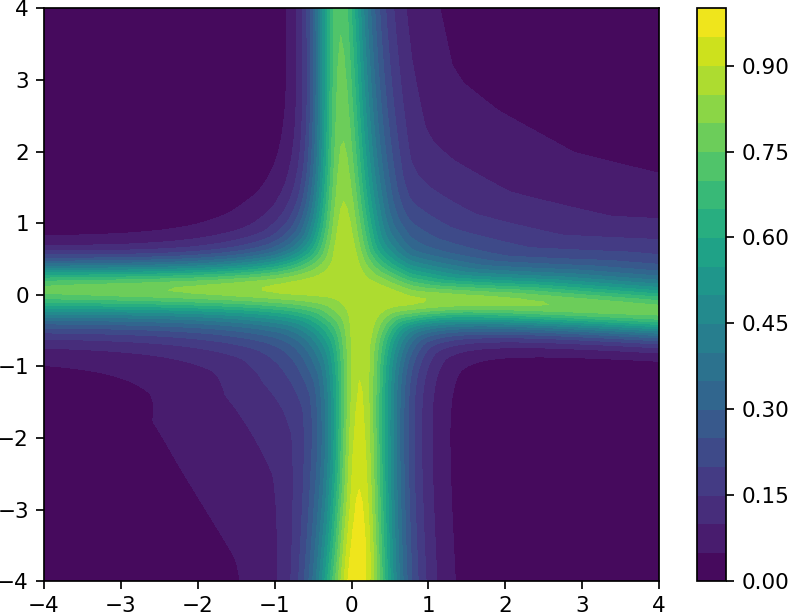}}}
    \,
    \subfloat[MVE]{{\includegraphics[width=4cm]{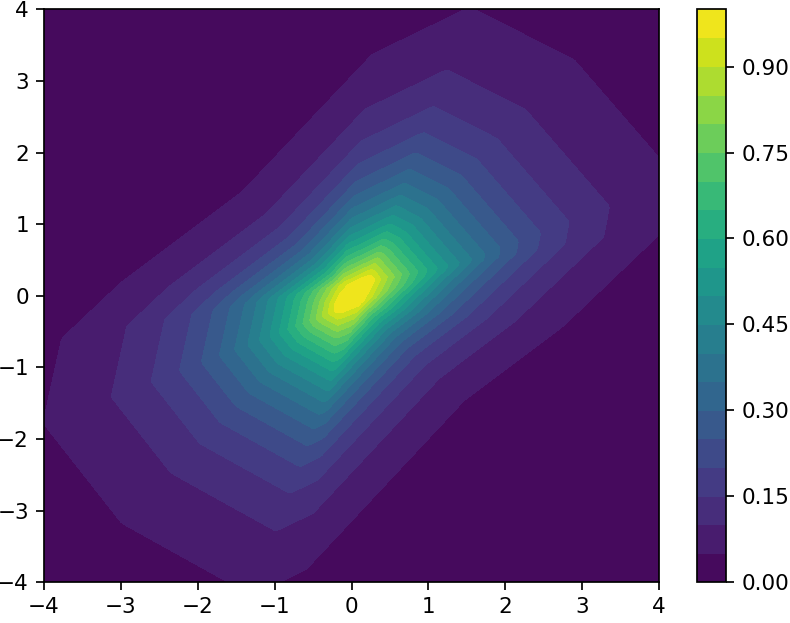}}}
    \,
    \subfloat[QD]{{\includegraphics[width=4cm]{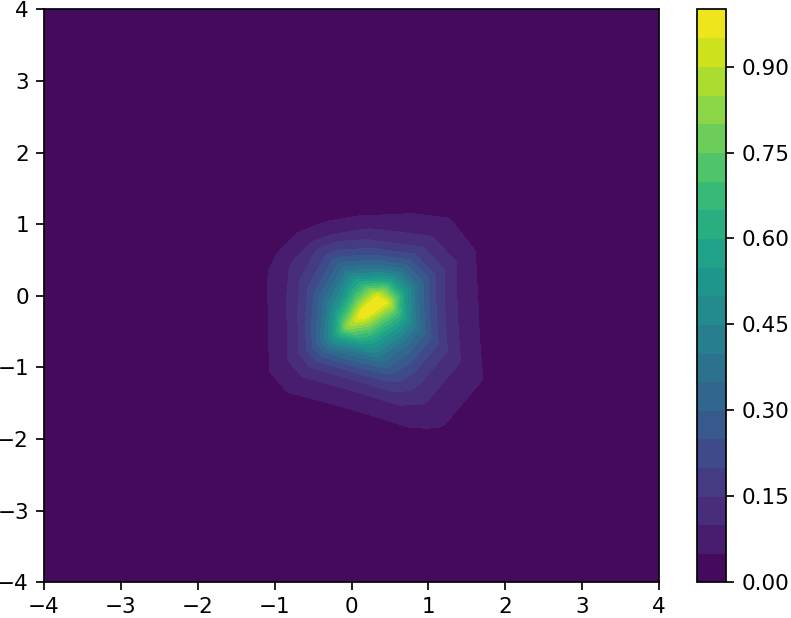}}}
    \,
    \subfloat[Confidence network]{{\includegraphics[width=4cm]{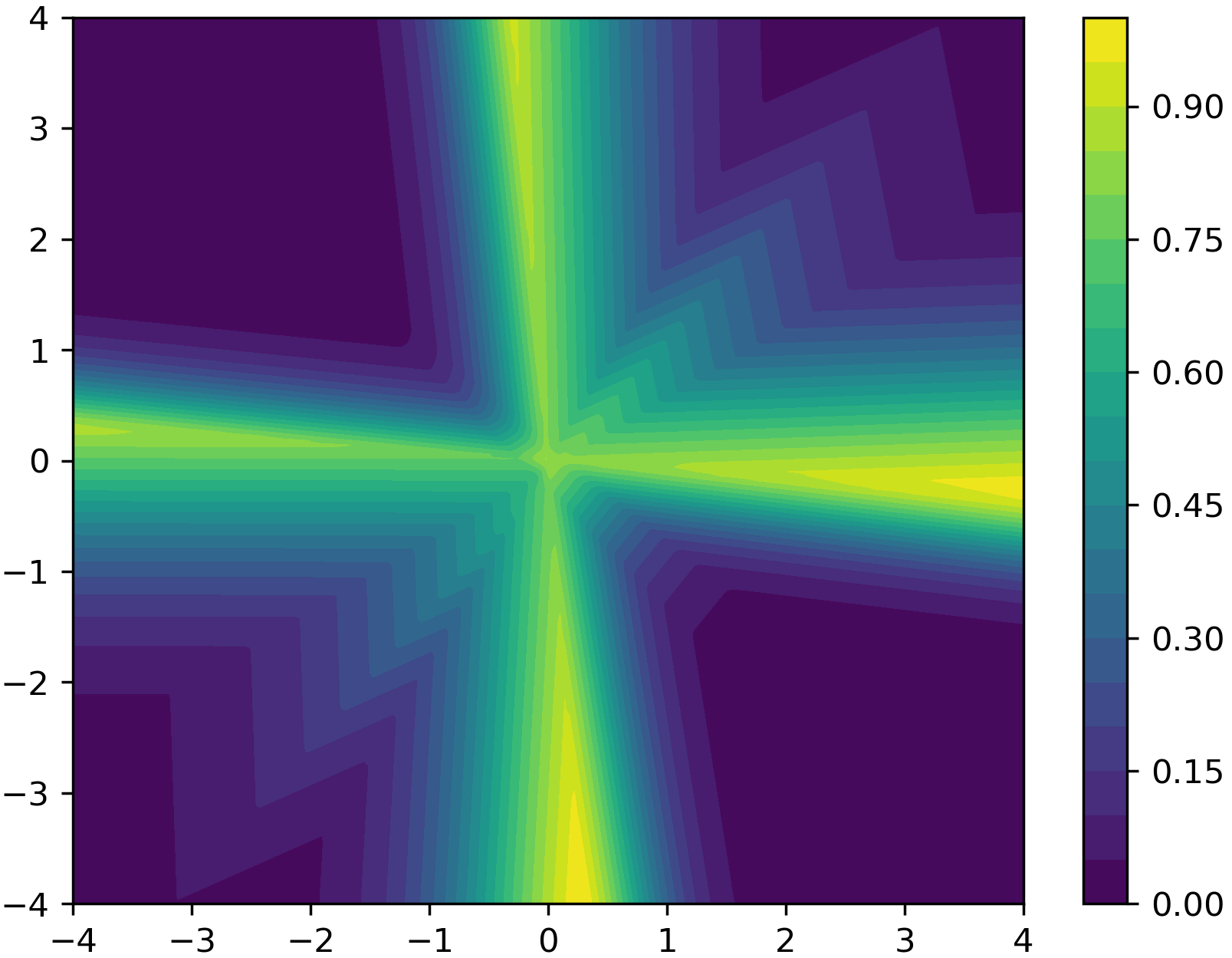}}}
    \,
    \caption{Variance contour plots of our approach and baselines. (a) shows the ideal variance plot. (b) is the result of our approach. (c), (d), (e) are the results of baselines. Blue means low data-noise, and yellow means high data-noise. From the results, (b) our approach looks most similar to the ideal.}
    \label{baselines}
\end{figure}

Figure \ref{baselines} reports the contour of the estimated variance. Panel (a) is the true variance contour for the simulated data, obtained numerically from the data generation. It shows that the largest variance occurs along the two axises that separate the four quadrants. Panel (b) is the result of our approach. We used ten mixtures here. The predicted mean and variance were calculated using the law of total expectation and total variance. Our method achieved a 98.4\% classification accuracy. More importantly, it successfully captured the variability of the classification probability and produced a variance contour that looks similar to (a). Panel (c) is the result of the mean variance estimation \cite{devries2018confidence}. It also achieved a 98.4\% classification accuracy, but it failed to correctly characterize the variability. This is partly due to that it models the variability as Gaussian. (d) is the result of the quality-driven prediction interval method \cite{pearce2018high}. It only obtained a 89.1\% classification accuracy. As a distribution-free method, it predicted a higher variability in the center, but ignored other highly variable regions. (e) is the result of the confidence network \cite{devries2018confidence}. It achieved a 98.1\% classification accuracy, a reasonably well variability estimation. Overall, our method achieved the best performance while maintaining a high classification accuracy.

\begin{figure}[h]
    \centering
    \subfloat[Point (0,0)]{{\includegraphics[width=4cm]{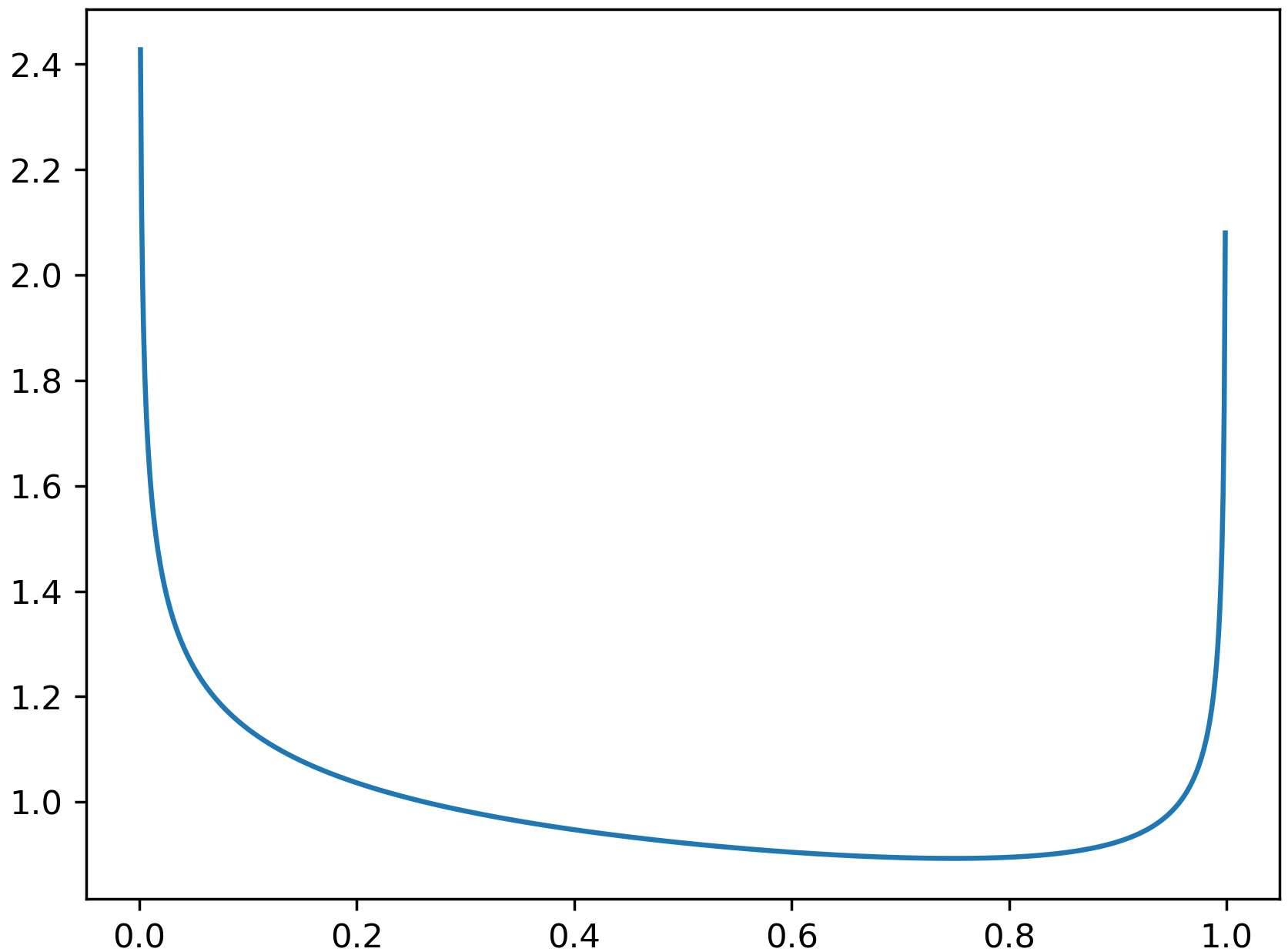}}}
    \,
    \subfloat[Point (1,1)]{{\includegraphics[width=4cm]{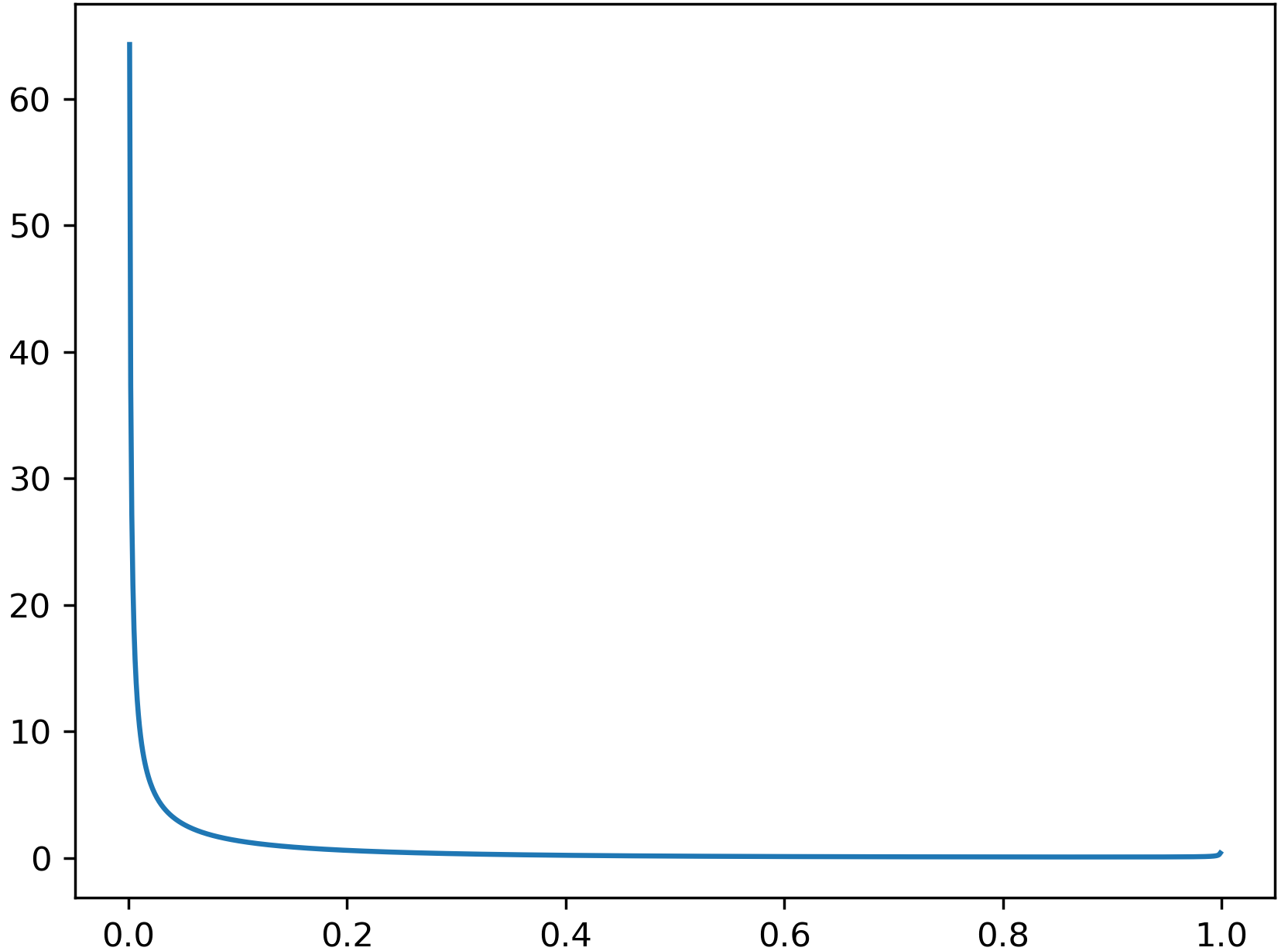}}}
    \caption{Beta mixture density functions outputted by the neural network. (a) is the result at point (0,0). (b) is the result at point (1,1). Point (0,0) clearly has a higher variance.}
    \label{PDFplots}
\end{figure}

Figure \ref{PDFplots} shows the density function of the outputted distributions. At point (0,0), it indeed has a higher variance.

\section{Real Data Analysis}
\label{sec:realdata}

\subsection{Data Description }

We illustrate our proposed method on a medical imaging diagnosis application. We remark that, although the example dataset is small in size, with only thousands of image scans, our method is equally applicable to both small and large datasets. 

Alzheimer's Disease (AD) is the leading form of dementia in elderly subjects, and is characterized by progressive and irreversible impairment of cognitive and memory functions. With the aging of the worldwide population, it has become an international imperative to understand, diagnose, and treat this disorder. The goal of the analysis is to diagnose patients with AD based on their anatomical magnetic resonance imaging (MRI) scans. Being able to provide an explicit uncertainty quantification for this classification task, which is potentially challenging and of a high-risk, is especially meaningful. The dataset we analyzed was obtained from the Alzheimer's Disease Neuroimaging Initiative (ADNI). For each patient, in addition to his or her diagnosis status as AD or normal control, two cognitive scores were also recorded. One is the Mini-Mental State Examination (MMSE) score, which examines orientation to time and place, immediate and delayed recall of three words, attention and calculation, language and vision-constructional functions. The other is the Global Clinical Dementia Rating (CDR-global) score, which is a combination of assessments of six domains, including memory, orientation, judgment and problem solving, community affairs, home and hobbies, and personal care. Although MMSE and CDR-global are not used directly for diagnosis, their values are strongly correlated with and carry crucial information about one's AD status. Therefore, we took the dichotomized cognitive scores, and used them as labels in addition to the diagnosis status. 

We used the ADNI 1-Year 1.5T dataset, with totally 1,660 images. We resized all the images to the dimension $96\times96\times80$. The diagnosis contains three classes: normal control (NC), mild cognitive impairment (MCI), and Alzheimer disease (AD). Among them, MCI is a prodromal stage of AD. Since the main motivation is to identify patients with AD, we combined NC and MCI as one class, referred as NC+MCI, and AD as the other class, and formulated the problem as a binary classification task. We used three types of assessments to obtain the three classification labels: the doctor's diagnostic assessment, the CDR-global score, and the MMSE score. For the CDR-global score, we used 0 for NC, 0.5 for MCI, and 1 for AD. For the MMSE score, we used 28-30 as NC, 24-27 as MCI, and 0-23 as AD. Table \ref{data} summarizes the number of patients in each class with respect to the three different assessments. 

\begin{table}[t]
\begin{center}
\begin{tabular}{l*{6}{c}r}
    & Diagnosis & CDR-global & MMSE \\ \hline
NC  & 500 & 664 & 785 \\
MCI & 822 & 830 & 570 \\
AD  & 338 & 166 & 305 \\ \hline
Total & 1660 & 1660 & 1660\\
\end{tabular}
\caption{Detailed patient statistics}
\label{data}
\end{center}
\end{table}

\subsection{Classifier and Results}

\begin{figure}[t!]
\centering
\includegraphics[width=8.2cm]{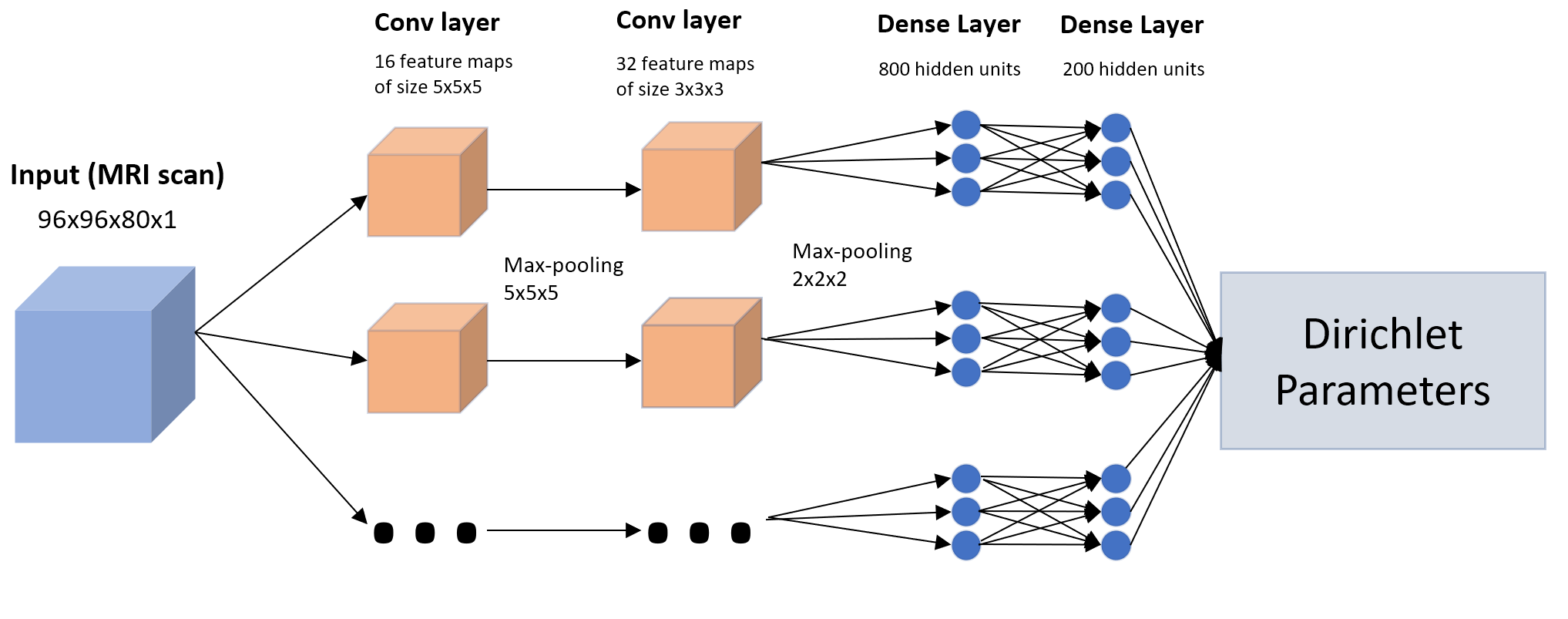}
\caption{Architecture of the neural network used in the real data experiment.}
\label{arch}
\end{figure}

Figure~\ref{arch} describes the architecture of our neural network based classifier. We used two consecutive 3D convolutional filters followed by max pooling layers. The input is $96\times96\times80$ image. The first convolutional kernel size is $ 5\times5\times5$, and the max pooling kernel is $5\times5\times5$. The second convolutional kernel is $3\times3\times3$, and the following max pooling kernel is $2\times2\times2$. We chose sixteen as the batch size, and $1e-6$ as the learning rate. We chose a $K=3$-component Beta mixture. 

\begin{figure}[t!]
\centering
\subfloat[Credible interval for 166 subjects]{{\includegraphics[width=4cm]{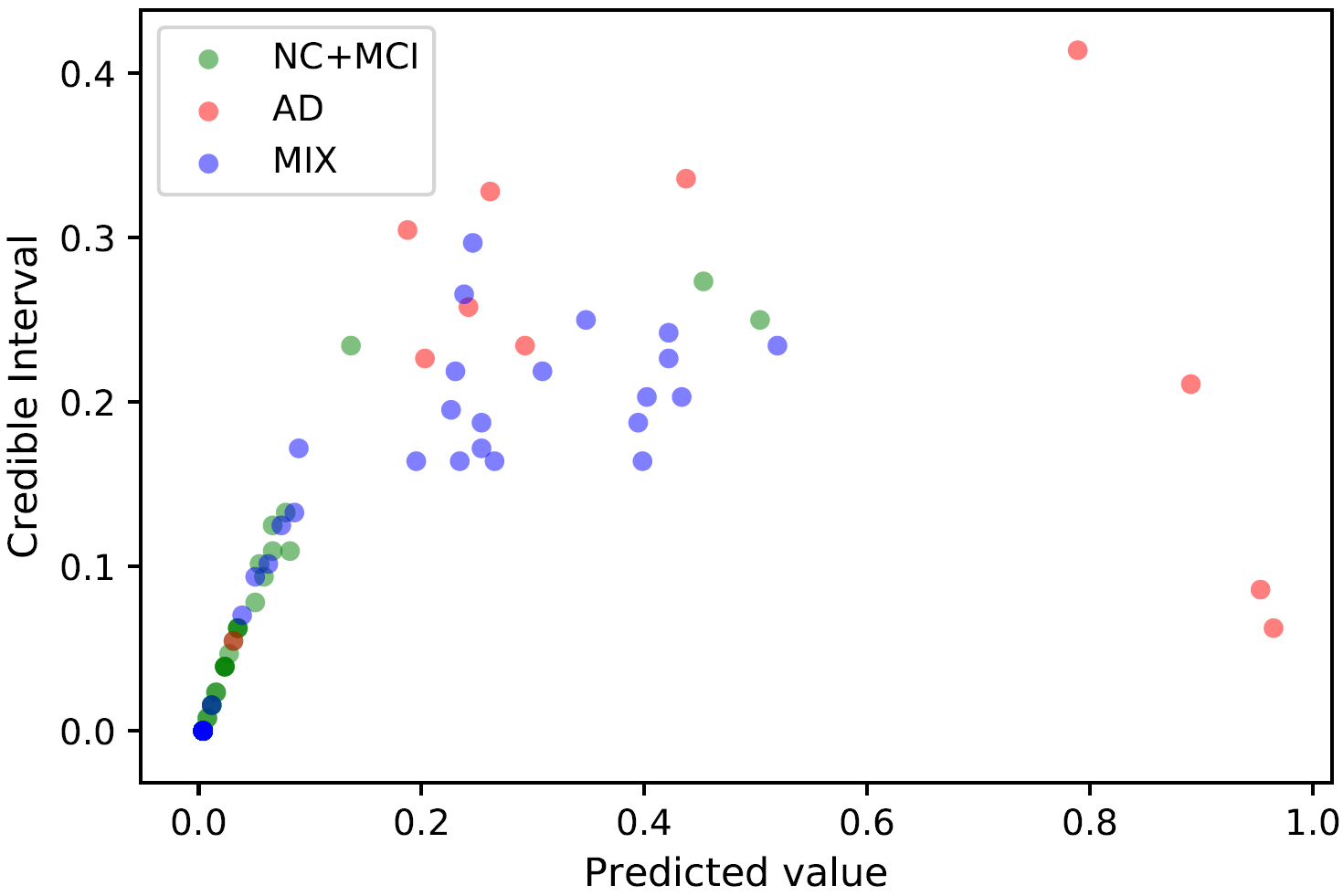}}}
\,
\subfloat[Patients with all NC+MCI label]{{\includegraphics[width=4cm]{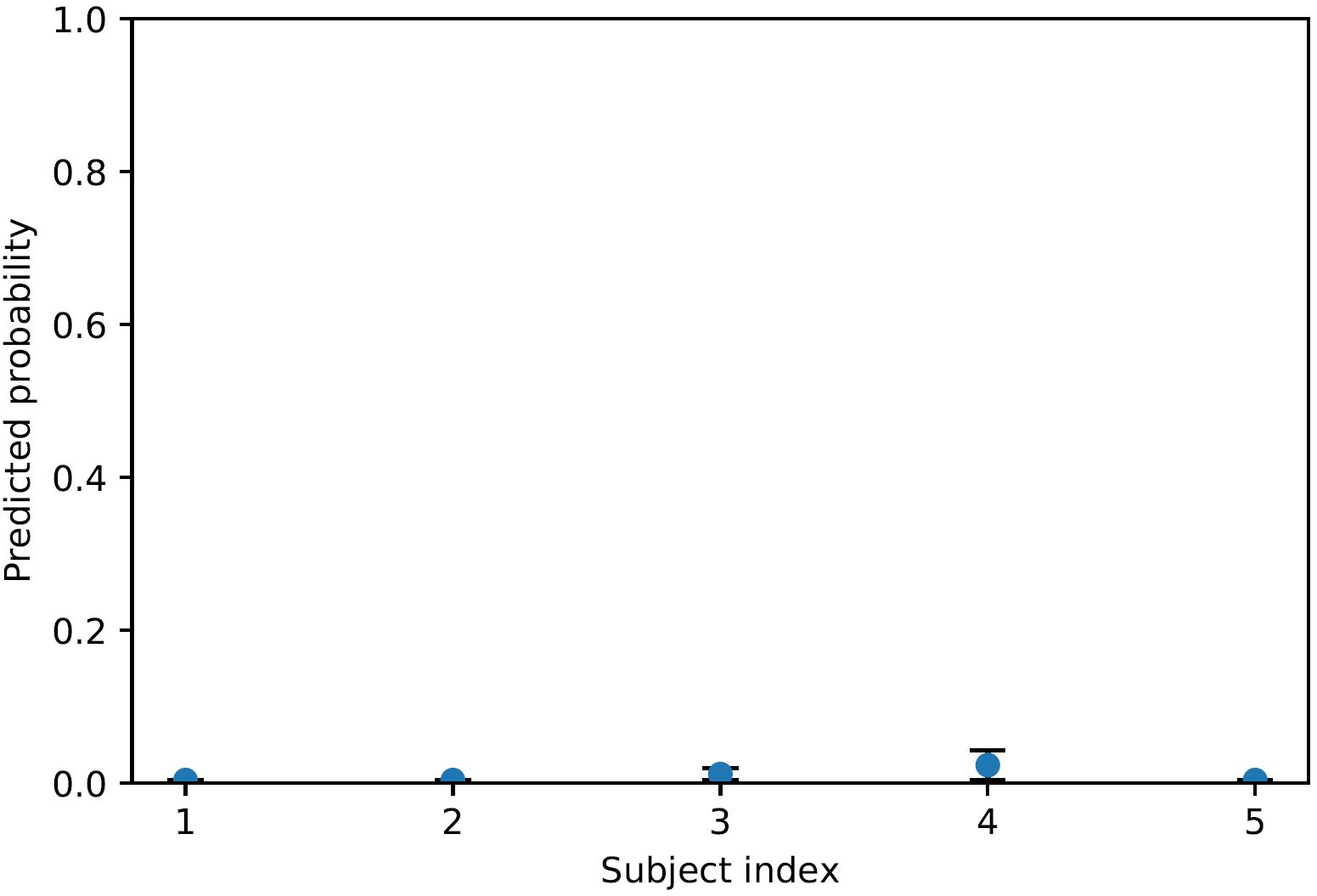}}}
\,
\subfloat[Patients with all  AD label]{{\includegraphics[width=4cm]{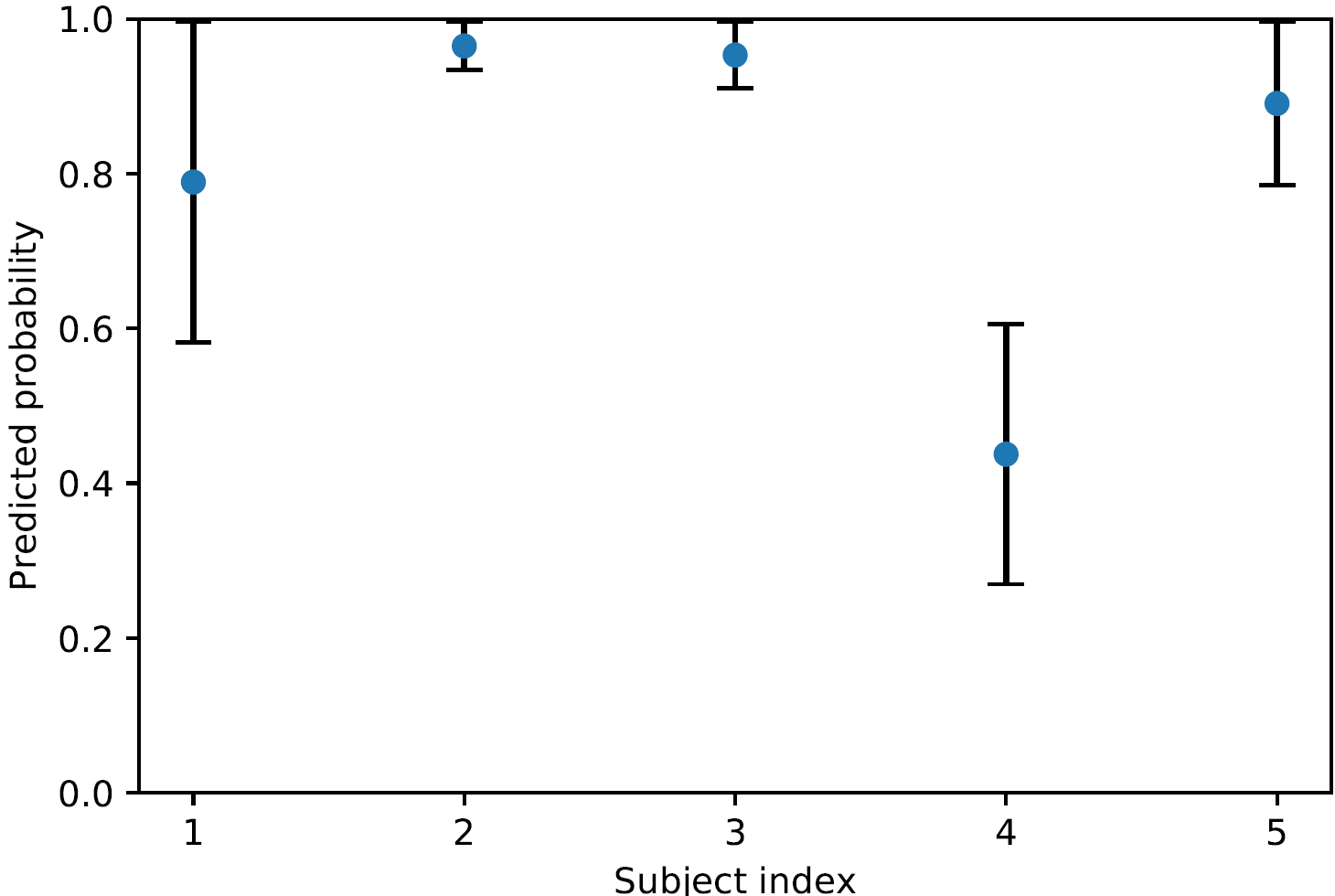}}}
\,
\subfloat[Patients with mixed AD and NC+MCI label]{{\includegraphics[width=4cm]{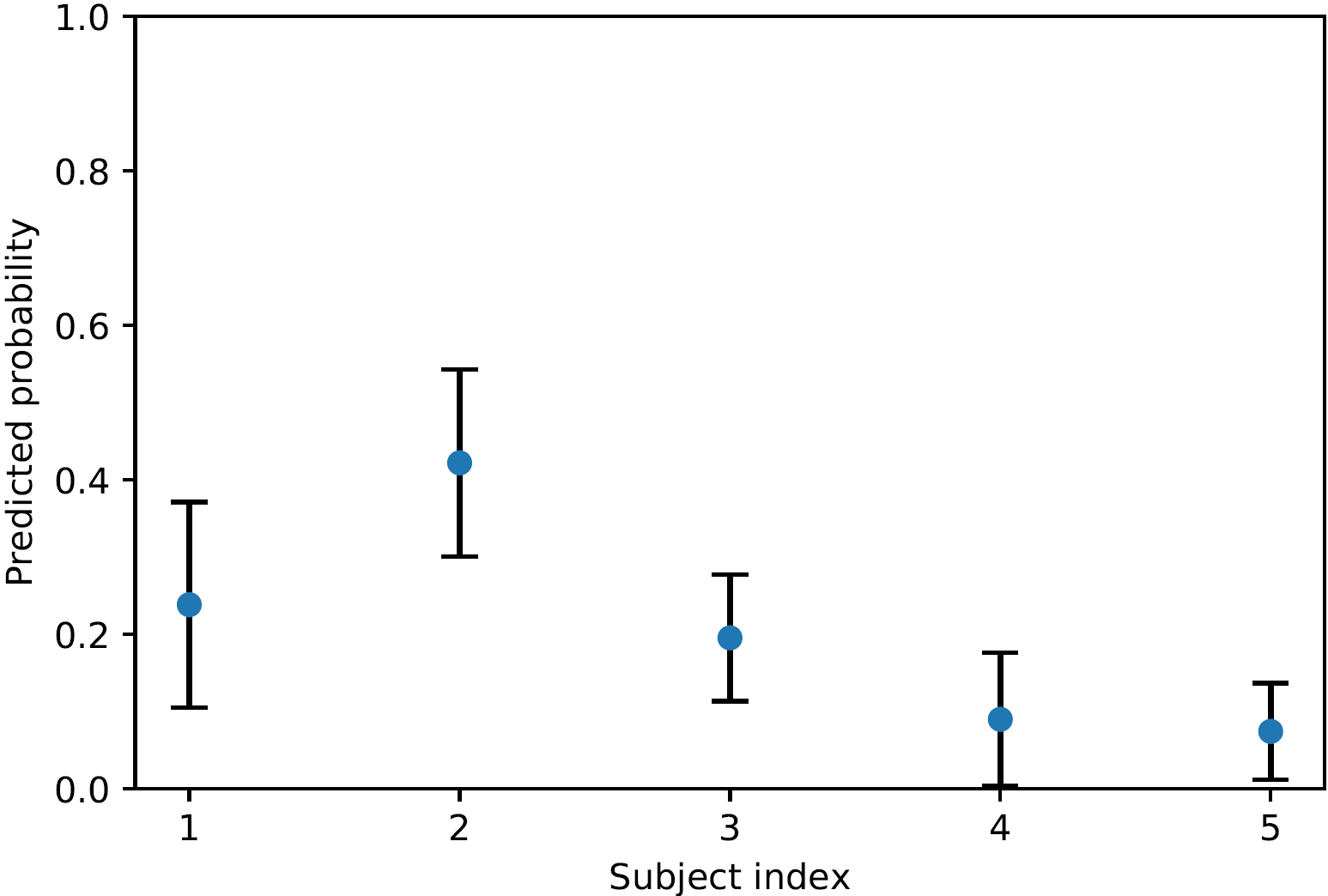}}}
\caption{Credible intervals constructed in the real data experiment.}
\label{real}
\end{figure}

We randomly selected 90\% of the data for training and the remaining 10\% for testing. We plotted the credible interval of all the 166 testing subjects with respect to their predicted probability of having AD or not in Figure \ref{real}(a). We then separated the testing data into three groups: the subjects with their assessments unanimously labeled as NC+MCI (green dots), the subjects with their assessments unanimously labeled as AD (red dots), and the subjects with their assessments with a mix of NC+MCI and AD (blue dots, and referred as MIX).

We observe that, for the patients in the NC+MCI category, 95\% of them were estimated to have a smaller than 0.1 probability of being AD, and a tight credible interval with the width smaller than 0.15. We further randomly selected five patients in the NC+MCI category and plotted their credible intervals in Figure \ref{real}(b). Each has a close to 0 probability of having AD, and each with a tight credible interval. For patients in the AD category, most exhibit the same pattern of having a tight credible interval, with a few potential outliers. For the patients in the MIX category, we randomly selected five patients and plotted their predicted classification probability with the associated credible interval in Figure~\ref{real}(c). We see that Subject 4 was classified as AD with only 0.45 probability but has a large credible interval of width 0.3. We took a closer look at this subject, and found that the wide interval may be due to inaccurate labeling. The threshold value we applied to dichotomize the MMSE score was 23, in that a subject with the MMSE below or equal to 23 is classified as AD. Subject 4 happens to be on the boundary line of 23. This explains why the classifier produced a wide credible interval. In Figure \ref{real}(a), we also observe that the classifier is less confident in classifying the patients in the MIX category, in that almost all the blue dots are above the 0.15 credible interval. We again randomly selected five patients in the MIX category and plotted their predicted classification probabilities with the corresponding credible intervals in Figure~\ref{real}(d). Comparing to Figure~\ref{real}(b), the credible intervals for patients in the MIX category are much wider than those in the unanimous NC+MCI category.

\section{Conclusion}

We present a new approach, deep Dirichlet mixture networks, to explicitly quantify the uncertainty of the classification probability produced by deep neural networks. Our approach, simple but effective, takes advantage of the availability of multiple class labels for the same input sample, which is common in numerous scientific applications. It provides a useful addition to the inferential machinery for deep neural networks based learning.

There remains several open questions for future investigation. Methodologically, we currently assume that multiple class labels for each observational sample are of the same quality. In practice, different sources of information may have different levels of accuracy. It is warranted to investigate how to take this into account in our approach. Theoretically, Petrone and Wasserman \cite{petrone2002consistency} obtained the convergence rate of the Bernstein polynomials. Our Dirichlet mixture distribution should at least have a comparable convergence rate. This rate can guide us theoretically on how many distributions in the mixture should we need. We leave these problems as our future research.

\newpage
\bibliography{reference.bib}
\bibliographystyle{icml2019}

\end{document}